\documentclass[runningheads]{llncs}
\usepackage[T1]{fontenc}
\usepackage{graphicx}
\usepackage{amsfonts}
\usepackage{makecell}
\usepackage{multirow}
\usepackage{float}
\usepackage{array}
\usepackage{amsmath}
\usepackage[utf8]{inputenc}  
\usepackage[T1]{fontenc}
\usepackage{placeins}
\begin{document}
\title{Long-Term Prediction of Local and Global Human Motion with Occlusion Recovery}
\titlerunning{Prediction of Local and Global Human Motion}
%
%
\author {Qiaoyue Yang\inst{1}\orcidID{0009-0009-8529-0652} \and 
	Sven Heutger\inst{1}\orcidID{0009-0009-1464-8602}\thanks{This work was done while S. Heutger was with Bielefeld University.} \and
	Christopher Niemann\inst{1}\orcidID{0009-0007-6122-1869} \and
Magnus Jung\inst{2}\orcidID{0000-0002-3632-2402} \and
Ayoub Al-Hamadi\inst{2}\orcidID{0000-0001-5371-7214} \and
Sven Wachsmuth\inst{1}\orcidID{0009-0002-2050-2967}}
\authorrunning{Q. Yang et al.}
%
\institute{COSY@Home-Lab, Faculty of Technology, Bielefeld University, Bielefeld, 
	33615, Germany \email{{qyang,cniemann,swachsmuth}@techfak.uni-bielefeld.de, sven.heutger@gmail.com} \and
Neuro-Information Technology, Otto-von-Guericke-University Magdeburg, Magdeburg, 
39106, Germany \email{{magnus.jung,ayoub.al-hamadi}@ovgu.de}
	}
%
\maketitle              
\begin{abstract}
Human motion describes the three-dimensional full-body movement of a person. Anticipating such motion holds significant relevance across a wide range of application domains such as human-robot interaction, autonomous driving, animation, and healthcare. In recent research, spatial and temporal dependencies are modeled by bidirectional attention mechanisms. These typically anticipate human motion in an autoregressive manner which could cause an accumulation of errors over time. As a consequence, they solely focus on local pose forecasting. To address these limitations, we propose a non-autoregressive transformer based on spatio-temporal attention, and train it not only for local pose anticipation, but also for global motion prediction in space. Furthermore, to enhance its applicability in real-world scenarios, our model is also trained to recover missing joints due to occlusions, and is capable of processing varying lengths of history observations. Our code is publicly available at \url{https://github.com/Q-Y-Yang/Prediction-of-Local-and-Global-Human-Motion}

\end{abstract}
\section{Introduction}
\label{sec:intro}
In today's era of intelligence, smart devices will no longer be confined to isolated environments. Instead, they will coexist within human environments, actively participating in daily life to help improve quality of life and work efficiency. Service robots and autonomous vehicles are among the first to be expected to assist humans in a human-like manner, building on the rapid evolution of AI technologies, particularly the success of large foundation models \cite{2024embodied,2024foundation}. Understanding human behavior is essential for enabling natural interactions with humans, and forecasting the human partner's motion  allows the robot to proactively plan its own interactive motion \cite{2020trajreview,24hmpreview}. Collaborative scenarios encompass stationary tasks that involve physical interaction, such as assembly operations at a production workstation, and dynamic tasks that require coordinated movement through space, as seen in warehouse logistics or mobile service robotics.

In this work, we address the anticipation of both local pose, that focuses on joint configurations within each pose frame without global position information, and global motion in space, that retains the global positional context. From the robot's ego-centric perspective, which provides only a single viewpoint, the human body is often partially occluded by other objects present in the environment. Therefore, our model is also trained to be robust to missing joints in the past poses observed, so that partial occlusions in the past frames do not significantly affect the prediction of future motion. Additionally, we exploit the Transformer-based models’ inherent capability to process sequential data by experimenting with the use of different lengths of past observations for training. This enables our approach to adaptively handle situations in real-world applications where a full history is not available or would negatively influence the prediction result.

The remainder of this paper is organized as follows: Section 2 reviews related work with an emphasis on 3D pose prediction. Section 3 presents a detailed description of our methodology along with the training setup. Section 4 provides both quantitative and qualitative evaluations, including ablation studies. Finally, Section 5 discusses and concludes this work.

\section{Related Work}
In the field of human motion prediction, the majority of studies have concentrated on 2D trajectory prediction and 3D pose forecasting. The former aims to predict a sequence of future waypoints, and the latter concentrates on the anticipation of fine-grained motion of human joints.

{\bf 2D Trajectory Prediction.} The past trajectory, along with contextual information such as a semantic map, is typically used to learn motion patterns. Some works also plan the future trajectory towards a predicted or known goal, as human behavior is usually goal-directed \cite{2020trajreview}. S2TNet \cite{s2tnet} models not only the temporal dependencies in traffic agents' past trajectories, but also the spatial interactions among different traffic agents. Bird's-eye view data is commonly utilized in this task, such as ETH \cite{eth}, UCY \cite{ucy}, and the Stanford Drone Dataset \cite{sdd}. Human Scene Transformer (HST) \cite{HST} is a multi-modal approach, leveraging not only past human positions, but also the 3D skeletal pose, head orientation and the scene to forecast human trajectories.

{\bf 3D Pose Forecasting.} The deterministic human motion prediction utilizes Recurrent Neural Networks (RNNs) \cite{seq2seq,graph} due to RNNs's capability of dealing with sequential time-series data \cite{RNN}, and its extensions such as Long Short-Term Memory (LSTM) and Gated Recurrent Unit (GRU) \cite{socialcontext,GRU,pvred,MSR-GCN}. Nevertheless, errors accumulate over time because of the auto-regressive design of RNNs \cite{transformer}, which leads to failures in the long-term prediction. 

In order to leverage prior knowledge of the human skeleton structure, \cite{MSR-GCN,GCNcui,traj} employ Graph Convolutional Neural Networks (GCN) by representing naturally connected joints (kinematic links of the human skeleton) as a connected graph to learn motions of all joints. Aksan et al. \cite{transformer} extend the Transformer architecture by leveraging temporal and spatial attention to learn the spatial relationships between joints and the temporal dependencies across pose frames, but it still generates prediction in an autoregressive manner. POTR \cite{port} and SPOTR \cite{sport} forecast future poses in a non-autoregressive way, predicting the entire future pose sequence in a single forward pass through the network. POTR \cite{port} does not use a mask in the decoder attention operations, allowing the decoder to freely model dependencies among elements in the output sequence. SPOTR \cite{sport} formulates human motion prediction as a sequence-to-sequence problem, and deploys graph convolutions alongside with self-attention mechanisms. In comparison with these methods, our approach captures per-joint spatio-temporal dependencies and generates predictions non-autoregressively.

\section{Method}
Given a pose sequence from time step 1 to $N$, denoted as $X_{1: N}  \in \mathbb R^{N \times J \times D} $, where J is the number of joints, and D is the dimension of each joint, the objective is to predict a future sequence from time step $N+1$ to $N+M$,
$	X_{N+1: N+M} \in \mathbb R^{M \times J \times D}$. 
Therefore, the task of human motion prediction is constructing a model $\mathcal{F}$ that 
\begin{equation}
	\widehat{X}_{N+1: N+M}=\mathcal{F}
	\left(X_{1: N}\right) \ 
\end{equation}
Typically, a prediction horizon of up to $1000 \ ms$ is defined as long-term, whereas a horizon up to $400 \ ms$ is referred to as short-term. Thus, under the condition of 30 $FPS$ (Frame per Second), the long-term prediction horizon is set to  $M=30$.

\subsection{Model Architecture}
{\bf Pose Embedding.} Inspired by Pose Transformers \cite{port}, we embed the pose sequence not only using a single Joint-separate linear layer, but also through a dedicated spatial attention layer that further processes the embedded poses. For each joint $j \in J$ at time step $t \in T$, firstly applying its own linear transformation:
\begin{equation}
	X_{t,j} = W^{(j)} \cdot X_{t,j,:}^\top + b^{(j)}
\end{equation}
where the input $X_{t,j,:}^\top \in \mathbb R^{D}$, $W^{(j)} \in \mathbb R^{{d_x} \times D}$, $b^{(j)} \in \mathbb R^{d_x}$, and the output $X_{t, j} \in \mathbb R^{d_x}$, $d_x$ refers to the defined joint embedding dimension. Notably, in order to recover the occluded joints, we replace the missing joints with learnable tokens. The spatial attention is then also computed separately for each joint.


{\bf Joint-separate Temporal Positional Encoding.} The joint feature vectors are embedded separately by the spatial attention of the pose embedding, which encodes the spatial configuration of joints in the body structure. However, temporal positional encoding is still required to represent the positional relationship between pose frames in a sequence. Therefore, we applied temporal Positional Encoding on each joint feature vector separately.

Our positional encoding is not applied on the full pose, otherwise the frequency
distributions vary between the feature vectors of different joints. The joints with
lower index have more high frequency positional information and joints with higher index have more lower frequency positional information, because the frequency of the positional encoding is dependent on the feature dimension. Therefore, applying positional encoding on every joint allows to capture joint positions locally (short-range dependencies) and globally (long-range dependencies).

\begin{figure}[H]
	\begin{center}
		\fbox{\includegraphics[width=\textwidth]{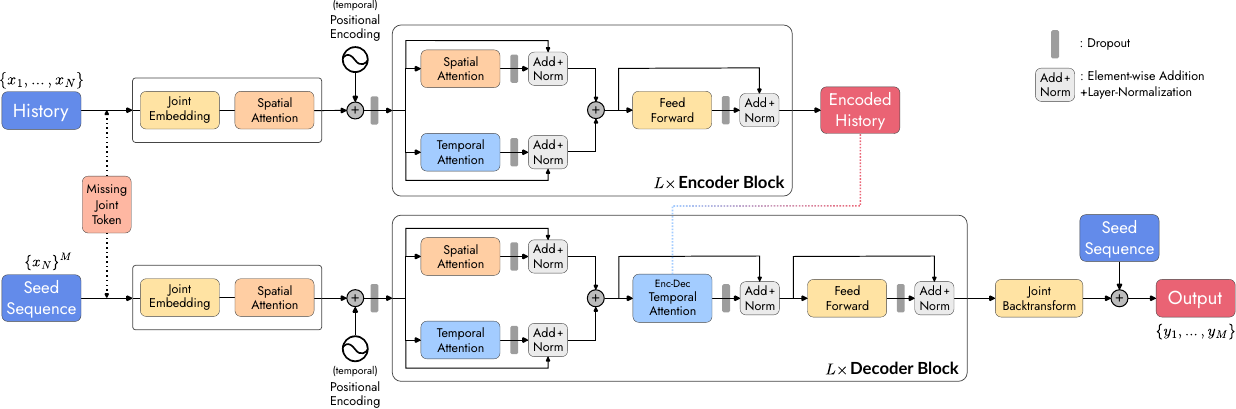}}
	\end{center}
	\caption{Non-Autoregressive Spatio-Temporal Transformer Architecture. The past pose sequence for the encoder and the seed sequence for the decoder are first embedded and then processed by spatial attention operations, respectively. After applying temporal positional encoding, the past sequence is passed through the encoder network (left), while the seed sequence is processed by the decoder network (right). The embedded joint outputs from the decoder are then transformed back into the original feature space and added as offsets to the seed sequence, forming the final predicted pose sequence.}
	\label{arcitecture}
\end{figure}

{\bf Spatio-Temporal Attention.} The spatio-temporal attention mechanism learns the spatial relationships among joints within each frame and the temporal dependencies of each individual joint across different time steps respectively. The spatial attention operates on each joint $j$ individually at every time step $t$. The query vectors $Q$ distinguishes between joints, whereas the parameters of the key $K$ and value vectors $V$ are shared across all joints,

\begin{equation}
	\label{form:spatial-attention-query}
	\begin{aligned}
		Q_t^{(h)} = \left[X_{t, 1}W^{(1, Q, h)}, .. , X_{t, J}W^{(J, Q, h)} \right], 
	\end{aligned}
\end{equation}
\begin{equation}
	\begin{aligned}
		K_t^{(h)} & = X_t W^{(K, h)},  \\
		V_t^{(h)} & = X_t W^{(V, h)}.
	\end{aligned}
\end{equation}

where $ Q_t^{(h)} \in \mathbb R^{J \times d_x}$, $ K_t^{(h)} \in \mathbb R^{J \times d_x}$, $ V_t^{(h)} \in \mathbb R^{J \times d_x}$, and $h$ denotes the index of the attention head. 
The multi-head attentions are then concatenated to obtain joint feature vectors $\hat X_t$,
\begin{equation}
	\begin{aligned}
		\text{head}_t^{(h)} = \text{Attention}(Q_t^{(h)}, K_t^{(h)}, V_t^{(h)}) \\
		\hat X_t = \text{Concat}(\text{head}^{(1)}_t, \dots, \text{head}_{ t}^{(H)}) W^{(B)}
	\end{aligned}
\end{equation}



The temporal attention is applied on each joint $j$, taking into account all of its timesteps ($1 : N$ for the history sequence, and $N+1 : N+M$ for the future sequence),
\begin{equation}
	\begin{aligned}
		Q^{(h)}_j & = X_{j} W^{(Q, h)}_j, 
		K^{(h)}_j & = X_{j} W^{(K, h)}_j, 
		V^{(h)}_j & = X_{j} W^{(V, h)}_j.
	\end{aligned}
\end{equation}
where $ Q^{(h)}_j \in \mathbb R^{T \times d_x}, K^{(h)}_j \in \mathbb R^{T \times d_x},  V^{(h)}_j \in \mathbb R^{T \times d_x} $. Separate weights are learned for each joint. In addition, the temporal attention is not masked, as our model is used in a non-autoregressive fashion, unlike the Spatio-Temporal Transformer \cite{transformer}.
The multi-head attentions are then concatenated to obtain joint feature vectors $\hat X_{j}$,
\begin{equation}
	\begin{aligned}
		\text{head}_j^{(h)} = \text{Attention}(Q^{(h)}_j, K^{(h)}_j, V^{(h)}_j) \\
		\hat X_j = \text{Concat}(\text{head}^1, \dots, \text{head}^H) W^{(B)}_j
	\end{aligned}
\end{equation}

{\bf Encoder}
Figure \ref{arcitecture} (left) depicts the overall architecture of the encoder. The sequence of observation history $X_{1:N}$ is fed to the encoder, in which poses are firstly embedded through the Pose Embedding and the temporal Positional Encoding. Next, the embedded poses are fed into the spatial attention and temporal attention blocks in parallel. The resulting outputs are summed and subsequently passed through a joint-separate position-wise feed-forward layer, ultimately yielding the encoded history.

{\bf Decoder}
As illustrated on the right side of Figure \ref{arcitecture}, our decoder generates the sequence of future predictions in a non-autoregressive manner. We copy the last pose of the history sequence to create the input seed sequence ${\left\{X_N\right\}}^{M}$ to the decoder. The decoder is constructed in similar fashion to the encoder, but adding an encoder-decoder temporal attention that retrospectively uses the encoded keys and values. We do not adopt an encoder-decoder spatial attention because the spatial attention should not process query and key-value pairs from different timesteps. Finally, the poses are projected back into 3D space using the Joint Backtransform layer, which is implemented as a joint-separate linear layer.

\subsection{Training} \label{train}

{\bf Pose Representation.} Joint poses effectively represent human motion by describing the spatial configuration of the human skeleton. Joint poses are primarily described using positional and rotational representations. Positional representation describes the 3D locations of joints in space, while rotational representation characterizes the orientation of human kinematic links using rotation matrices.

{\bf Joint Occlusion.} We simulate joint occlusions during training by masking joints in the data, in order to improve the model's robustness to occlusions. Three types of occlusions are employed to simulate real-world occlusion scenarios: (1) individual joints, (2) multiple adjacent joints, and (3) temporal masking. Specifically, individual joint masking randomly occludes a single joint within a frame, whereas multiple adjacent joints are masked together by randomly selecting a region from a predefined set, including the lower arms, legs, head, hands, and feet. Lastly, temporal masking involves occluding the same joint or the same set of adjacent joints across several consecutive frames (3-10 frames), requiring the model to exploit temporal continuity for accurate recovery.

{\bf Sequence Construction.} The motion clips from the full dataset are first randomly divided into training, validation, and test sets. The AMASS dataset \cite{AMASS} is split by subsets, and Human3.6M \cite{human3.6m} and HA4M \cite{ha4m} are partitioned by subject. The frames within each clip are further segmented into sequences, which serve as input samples. To reduce redundancy from highly overlapping samples while maintaining comprehensive coverage of the motion patterns within the dataset, we adopt a sliding window approach with random starting points. We employ a sliding window of 0.5 seconds and randomly select a starting frame within each window.

{\bf Training Setup.} Our model was trained with a standard $L1$ loss for both the positional and joint-angle representations, because the predicted poses tend to converge to a mean pose if using $L2$ loss during training \cite{GAN}. Given a predicted future motion sequence $\hat Y = \{ \hat y_1, \dots \hat y_M\}$ and its corresponding groundtruth sequence $Y = \{ y_1, \dots, y_M \}$, the $L1$ loss function is defined as:
\begin{equation}
	\begin{aligned}
		\mathcal L_{L1} = \frac{1}{MJ} \sum\limits_{t=1}^M \sum\limits_{j=1}^J || \hat y_t^{(j)} - y_t^{(j)} ||_1
	\end{aligned}
\end{equation}

The model was trained for a maximum of 100 epochs with early stopping. Early stopping terminates the training process if the validation loss fails to improve over a specified number of consecutive epochs. This prevents the network from overfitting to the training data. We used AdamW \cite{adamw} optimizer with a learning rate of $10^{-04}$ and weight decay of $10^{-05}$. Further, the batch size was chosen as 128 training samples.

\section{Experiments}

Our model was trained on three public datasets. Human3.6M \cite{human3.6m} is a common public benchmark dataset. Its training data contains 3D human poses and corresponding images of 7 actors in 15 scenarios (directions, discussion, sitting on chair, smoking, making purchases, greeting, waiting, taking photo, talking on the phone, walking, walking dog, walking together, posing, eating, activities while seated). The archive of motion capture as surface shapes (AMASS) \cite{AMASS} is a collection of 24 motion capture datasets, ranging from a variety of daily movements to more specialized actions such as dancing. The HA4M \cite{ha4m} dataset contains motion sequences of individuals performing a gear train assembly task. The recordings were captured using a single Azure Kinect camera.
\FloatBarrier
\subsection{Quantitative Results}
{\bf Evaluation Metrics} Mean Per Joint Position Error (MPJPE) and the Mean Angle Error (MAE) are common metrics to evaluate the precision of the anticipated poses.  The MPJPE measures the Euclidean distance between the predicted 3D joint locations and their corresponding groundtruth locations.
The MAE measures the Euclidean distance between the predicted and ground-truth Euler angles for each joint.
The set of Euler angles represents the relative joint orientations of the corresponding joint.

We evaluated the MPJPE at six time steps over a span of 1000 milliseconds for the three datasets shown in Tables \ref{mae-h36m}, \ref{mae-amass}, and \ref{mae-ha4m}, respectively. The results show that our method outperforms a baseline implementation using the standard Transformer architecture \cite{allyouneed}. Especially, the occlusion model (trained and tested with missing joints in the observation history, as described in Section \ref{train}) achieves comparable performance to that of the model trained and tested on fully observed data. It indicates strong recovery capabilities from the spatial and temporal relationships. Furthermore, the reported values under varying input lengths represent the test results at each time step for $1$-second future predictions, averaged over observation windows of $0.75, 1, 1.5,$ and $1.75$ seconds. During both training and testing, the history window length is randomly sampled from a uniform distribution ranging from 0.75 to 1.75 seconds.\\

\begin{table}[!htbp]
	\begin{center}
		\begin{tabular}{|l|cccccc|cccccc|}
			\hline
			Method & \multicolumn{6}{c|}{local MPJPE at Timestep (ms)}&\multicolumn{6}{c|}{global MPJPE at Timestep (ms)} \\
			\textbf{} & 80 & 160 & 320 & 400 & 560 & 1000& 80 & 160 & 320 & 400 & 560 & 1000 \\
			\hline\hline
			Transformer&25.1&41.6&65.9&83.4&98.7&143.7&24.5& 42.8& 75.5& 90.7& 120.3& 201.9\\
			ours & 17.2 &31.1& 55.6& 68.7& 84.6& 123.9 &19.0& 34.0& 62.0& 75.3& 101.3& 173.7\\
			ours (occlu.)&17.6&31.4&55.6&66.3&84.6&123.6&19.0&34.4&63.5&77.2&103.5&176.3 \\
			ours (vary.)&20.5&35.6&60.8&71.4&89.1&127.3&21.2&38.2&69.8&84.8&106.6&193.6 \\
			\hline
		\end{tabular}
	\end{center}
	\caption{Local and Global MPJPE (in millimeters) on the Human3.6M dataset (lower is better). (occlu.) represents training and testing with joint occlusions. (vary.) denotes the use of input sequences with varying lengths.} \label{mae-h36m}
\end{table}
\FloatBarrier
\begin{table}[!htbp]
	\begin{center}
		\begin{tabular}{|c|cccccc|cccccc|}
			\hline
			Method & \multicolumn{6}{c|}{local MPJPE at Timestep (ms)}&\multicolumn{6}{c|}{global MPJPE at Timestep (ms)} \\
			& 100 & 200 & 300 & 400 & 600 & 1000& 100 & 200 & 300 & 400 & 600 & 1000 \\
			\hline\hline
			Transformer&28.3&49.3&69.4&89.7&108.6&145.0&34.3 &60.4 &81.1& 98.7& 114.6& 199.4\\
			ours &19.3& 35.9 &50.4& 62.0& 79.2& 105.3 & 25.7 &48.3& 68.1& 86.2& 103.3& 186.8\\
			\makecell{ours (occlu.)}&19.6&36.7&51.3&63.1&81.1&106.0&25.9&48.2&67.7&85.3&116.3&188.2 \\
			ours (vary.)&22.3&40.1&55.1&66.9&84.1&109.8 &35.5&63.9&87.4&105.0&145.5&214.8\\
			\hline
		\end{tabular}
	\end{center}
	\caption{Local and Global MPJPE (in millimeters) on the AMASS dataset (lower is better). (occlu.) represents training with joint occlusions. (vary.) denotes the use of input sequences with varying lengths.}	\label{mae-amass}
\end{table}

\begin{table}[!htbp]
	\begin{center}
		\begin{tabular}{|c|ccccccc|}
			\hline
			Method & \multicolumn{7}{c|}{MPJPE at Timestep (ms)} \\
			& 100 & 200 & 300 & 400 & 600 & 1000 & Avg. \\
			\hline\hline
			Transformer&24.4& 38.4 &49.9& 59.7& 68.1 &95.8&65.4\\
			ours &22.2& 33.9& 43.5& 52.1& 59.9& 87.8&57.2\\
			ours (occlu.) & 21.9 & 33.5 & 43.0 & 51.4 & 59.0 & 87.2 & 56.8\\
			\hline
		\end{tabular}
	\end{center}
	\caption{Local MPJPE (in millimeters) on the HA4M dataset (lower is better). (occlu.) represents training with joint occlusions.}	\label{mae-ha4m}
\end{table}
\FloatBarrier
\begin{table}[!htbp]
	\begin{center}
		\begin{tabular}{|c|cccccc|}
			\hline
			Method & \multicolumn{6}{c|}{MAE at Timestep (ms)} \\
			& 80 & 160 & 320 & 400 & 560 & 1000 \\
			\hline\hline
			SPOTR \cite{sport} & 0.33 & 0.58& 0.89 &\underline{1.00} & - & - \\
			POTR \cite{port} & \textbf{0.22}&\underline{0.56}& 0.94& 1.01& - & -\\
			ST-Transformer \cite{transformer} &\underline{0.30}& \textbf{0.55}& \underline{0.90}& 1.02&-&-\\
			ours & 0.57&0.71&\textbf{0.88}&\textbf{0.96}&1.06&1.27\\
			\hline
		\end{tabular}
	\end{center}
	\caption{Comparison of Local MAE on the Human3.6M dataset with state-of-the-art methods (lower is better). } \label{sota}	
\end{table}
Moreover, we also compared the MAE at different time steps on Human3.6M with other state-of-the-art approaches based on spatio-temporal attention in Table \ref{sota}.  In the long-term, ours has lower errors than non-autoregressive SPOTR \cite{sport} and POTR \cite{port}, and the autoregressive ST-Transformer \cite{transformer}.

In addition, inference time is measured as the average over 1,000 runs. A single prediction takes $250 ms$ and $247 ms$ on the Tesla P100 and GTX 1080 Ti, respectively. At this scale of latency, the predicted future does not become obsolete by the time it is produced, since the prediction horizon is one second.\\

{\bf Ablation Studies} were conducted on Human3.6M for global motion prediction to facilliate the final design of the proposed network architectures, as listed in Table \ref{ablation}. First, the joint embedding projects 3D representation of joints into a higher dimensional space, and we found out that a dimensionality of 64 is superior than either a smaller or larger one. Second, it was evaluated whether the decoder should contain a second spatial attention. Without a second spatial attention outperforms with one either positioned in parallel to the encoder-decoder temporal attention or after it. Last, the position-wise feed-forward network of the transformer \cite{allyouneed} increases the dimension of joint features and then projects it back. We investigated the ratio between the dimensionalities of the joint feature embedding and the intermediate representation in the feed-forward network, and found that a ratio of $1:2 (64:128)$ is sufficient.

In addition, joint motion is influenced by parent joints in the kinematic chain, for example, the position of the wrist in space changes when the elbow or shoulder moves, even if the wrist itself does not move. We hypothesized that representing joints relative to their parent joints in the kinematic tree would improve prediction performance, given that it incorporates the kinematic structure. However, in our experiments, the absolute representation clearly performs better than the relative representation, as listed in Table \ref{relative}. This may be due to errors in the child joints accumulating over errors in the parent joints.

\begin{table}[!htbp]
	\begin{center}
	\scalebox{0.85}{
		\begin{tabular}{|c|c|ccccccc|}
			\hline
			\multicolumn{2}{|c|}{Method} & \multicolumn{7}{c|}{MPJPE at Timestep (ms)} \\
			\multicolumn{2}{|c|}{}& 80 & 160 & 320 & 400 & 560 & 1000& Avg. \\
			\hline\hline
			\multirow{4}{*}{\makecell{Embedding \\Dimensions}} &16 & 21.0 & 37.0 & 67.5 & 81.7 & 108.7 & 185.5 & 101.1\\
			&32 & \textbf{19.7} & \textbf{35.7} & 65.1 & 78.5 & 104.8 & 179.2 & 97.4\\
			& 64 & 20.5 & 35.8 & \textbf{64.4} & \textbf{77.7} & \textbf{103.7} & \textbf{178.5} & \textbf{96.9} \\
			& 128 & 26.0 & 42.0 & 73.1 & 87.6 & 115.9 & 195.5 & 108.2 \\
			\hline
			\multirow{3}{*}{	Decoder Block} & Parallel & 19.5 & 34.4 & 63.0 & 76.5 & 102.7 & 175.2 & 95.3\\
			& Sequential & 19.4 & 34.2 & 62.9 & 76.6 & 103.1 & 177.3 & 95.8 \\
			& Temporal-only & \textbf{19.1} & \textbf{33.9} & \textbf{61.5} & \textbf{74.8} & \textbf{100.5} & \textbf{172.4} & \textbf{93.4} \\
			\hline
			\multirow{3}{*}{\makecell{Embedding :\\Feed Forward}}	&  1:4 (64:256) & 23.5 & 40.8 & 55.6 & 68.1 & 79.0 & \textbf{112.0} & 67.9 \\
			& 1:2 (64:128) & \textbf{22.8} & \textbf{39.9} & \textbf{54.8} & \textbf{67.5} & \textbf{78.7} & 112.7 & \textbf{67.7} \\
			& 1:1 (64:64) & 23.8 & 40.9 & 55.8 & 68.5 & 79.7 & 114.2 & 68.7 \\
			\hline
		\end{tabular}}
	\end{center}
	\caption{Ablation Studies of the Model Architecture.} \label{ablation}
\end{table}
\FloatBarrier
\begin{table}[!htbp]
	\begin{center}
	\begin{tabular}{|c|ccccccc|}
		\hline Pose Representations & \multicolumn{7}{c|}{ MPJPE at Timestep (ms) } \\
		& 80 & 160 & 320 & 400 & 560 & 1000 & Avg. \\
		\hline \hline Rotation (Abs.) & 22.7 & 35.5 & 59.3 & 69.5 & 87.3 & 125.6 & 78.6 \\
		Rotation (Rel.) & 35.5 & 47.9 & 71.3 & 82.0 & 100.5 & 142.9 & 92.3 \\
		\hline
	\end{tabular}
\end{center}
	\caption{Local MPJPE (in millimeters) on the Human3.6M dataset (lower is better). (Abs.) denotes representation of each joint using an absolute rotation matrix, while (Rel.) indicates that each joint is represented by a rotation matrix relative to its parent joint.} \label{relative}
\end{table}
\vspace{-8mm}
\subsection{Qualitative Results}
\begin{figure}
	\begin{center}
		\fbox{\includegraphics[width=\textwidth]{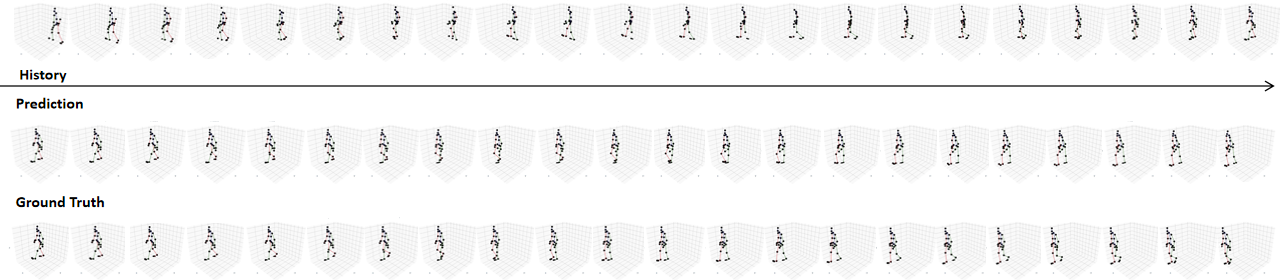}}
	\end{center}
	\caption{Visualization of global motion anticipation for walking and turning behaviors. The top row shows observation history, the middle row displays model predictions, and the bottom row presents the ground truth.}
	\label{globalwalking}
\end{figure}

Figure \ref{globalwalking} presents a global motion prediction when the person is walking and turning around simultaneously. The model is capable of anticipating both the global position change in the space and the walking pattern.

\begin{figure}
	\begin{center}
		\fbox{\includegraphics[width=0.85\textwidth]{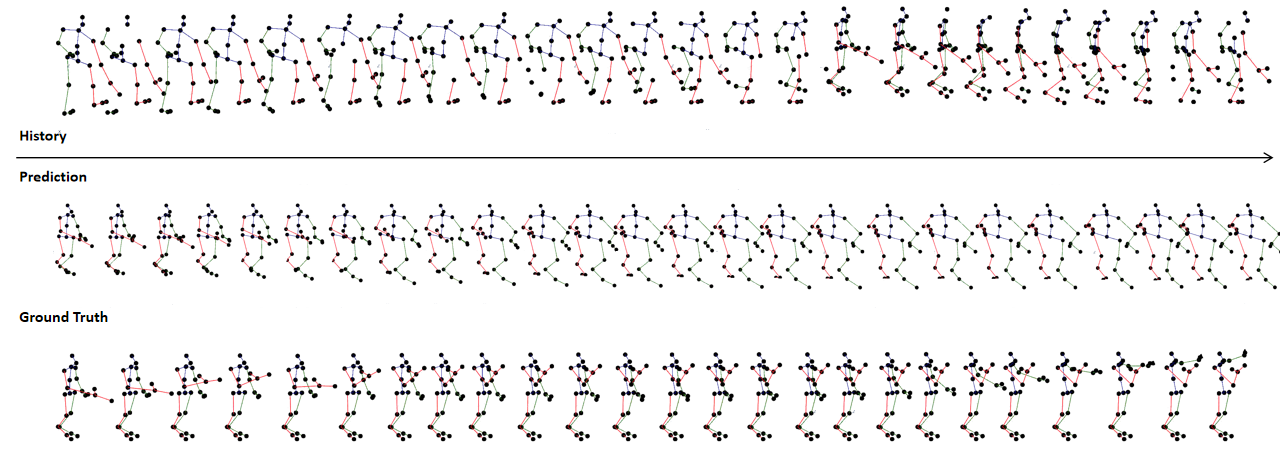}}
	\end{center}
	\caption{Visualization of local pose anticipation for a motion sequence that begins with walking and transitions into a stationary arm activity.}
	\label{failure}
\end{figure}

Figure \ref{failure} illustrates a typical failure case where the person is walking and turning, and then stops and starts to do something else with the hands. Since the observed history only sees the person is walking and turning, the anticipation is turning further. This type of data sequence covers the change of the human action, i.e. two different actions. In this case, the temporal relationship between history and future is very weak. Even the human brain is unable to foresee the next motion without overall context information when the action is changing.

\begin{figure}
	\begin{center}
		\fbox{\includegraphics[width=\textwidth]{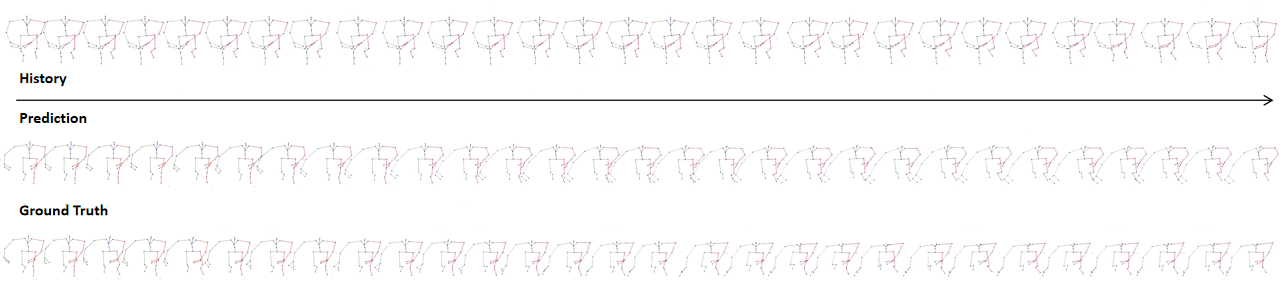}}
	\end{center}
	\caption{Visualization of local pose anticipation for an assembly task from HA4M dataset.}
	\label{ha4m}
\end{figure}

In addition, Figure \ref{ha4m} visualizes the result of an assembly task in the HA4M dataset, which focuses on arm motion at a workstation. The reaching motion is accurately anticipated based on the observation history that exhibits only the initial signs of arm extension

Based on our experimental observations, common behaviors such as walking and running, which are the most represented actions across datasets, can be predicted reliably. These motion patterns are also often implicitly present, even within action categories not explicitly labeled as walking or running. When global motion is accurately predicted, some highly complex body movements tend to be overlooked. Conversely, when the model focuses solely on local pose prediction, the accuracy of predicting body movements improves, though diverse arms' and hands' motions remain challenging to be captured precisely.
\FloatBarrier
\FloatBarrier

\vspace{-3mm}
\section{Discussion \& Conclusion}
Human motion sequences theoretically have infinite variations, even under kinematic constraints. Poor predictions can be attributed to two potential factors. First, data scarcity: if a type of motion has high variability but occurs infrequently in the training data, prediction quality tends to degrade, for example, in the case of complex and diverse arm movements. However, in the HA4M dataset, where the motion is concentrated in the arms and hands, the movements are primarily related to assembly tasks and exhibit less variability, leading to promising predictive performance. Second, the unpredictability of human motion transitions, as our failure case analysis shows. It is unreasonable to use only the observation history of the previous action to predict the motion of the next action. To achieve that, broader behavioral context is necessary.

In conclusion, we propose an adaptive human motion prediction framework capable of predicting both global and local motion. To reflect real-world scenarios, we also trained the model on joint-occluded training data. Compared to training without occlusion, the performances do not degrade. Our model offers the flexibility to handle motion histories of varying lengths, although the performance on 1-second future anticipation does not improve compared to using a fixed 1-second history. Our approach outperforms other non-autoregressive methods in the long-term prediction. For future work, we suggest either fine-tuning motion prediction for specific application scenarios to improve task relevance and accuracy, or expanding the model by incorporating action context information from task planning, and training on large-scale data to enhance generalization.

\begin{credits}
\subsubsection{\ackname} This work is funded by Deutsche Forschungsgemeinschaft (DFG, German Research Foundation) under grant number 502483052. C. Niemann is supported by the BMV DZM enableATO project (19DZ23002B).

\subsubsection{\discintname}
The authors have no competing interests to declare that are relevant to the content of this article. 
\end{credits}
%
%
%
%

\end{document}